\documentclass[11pt]{article}

\usepackage{booktabs}
\usepackage{amsmath}
\usepackage{amsfonts}
\usepackage{enumitem}
\usepackage{multirow}
\usepackage{algorithm}
\usepackage{algpseudocode}
\usepackage{acl}

\usepackage{times}
\usepackage{latexsym}

\usepackage[T1]{fontenc}

\usepackage[utf8]{inputenc}

\usepackage{microtype}

\usepackage{inconsolata}

\usepackage{graphicx}

\usepackage{tcolorbox}
\usepackage{caption}
\tcbuselibrary{skins,breakable}

\newtcolorbox{promptbox}[1]{
  colback=gray!5,
  colframe=black!75,
  colbacktitle=gray!40,
  coltitle=black,
  title=#1,
  fonttitle=\bfseries\fontsize{8.2pt}{9.3pt}\selectfont,
  fontupper=\fontsize{8.2pt}{9.3pt}\selectfont,
  breakable,
  enhanced,
  boxrule=0.5pt,
  arc=0pt,
  outer arc=0pt,
  left=3pt, right=3pt, top=3pt, bottom=3pt
}

\title{ CreaMem: A Scene-Aware Memory Architecture for Personalized Agents}

\author{
  \textbf{Qixuan Sun\textsuperscript{1,3}},
  \textbf{Yue Que\textsuperscript{1}},
  \textbf{Bowei He\textsuperscript{2}},
  \textbf{Jin Guo\textsuperscript{3,*}},
\\
  \textbf{Dihang Yang\textsuperscript{3}},
  \textbf{Wenchang Situ\textsuperscript{3}},
  \textbf{Chen Ma\textsuperscript{1,*}}
\\
  \textsuperscript{1}City University of Hong Kong, Hong Kong SAR, China
\\
  \textsuperscript{2}Mohamed bin Zayed University of Artificial Intelligence, Abu Dhabi, UAE
\\
  \textsuperscript{3}AgentWoods Inc., San Francisco
\\
  \small{
    \texttt{\{qixuansun2-c,yueque2-c\}@my.cityu.edu.hk},
    \texttt{Bowei.He@mbzuai.ac.ae}
  }
\\
  \small{
    \texttt{\{jin,dihang,wenchang.situ\}@creaseed.ai},
    \texttt{chenma@cityu.edu.hk}
  }
\\
  \small{\textsuperscript{*}\textbf{Corresponding authors}}
}

\begin{document}
\maketitle
\begin{abstract}
Long-term memory is a core capability for personalized LLM agents. 
To support it, existing memory systems organize information using various criteria such as topic segments or summary hierarchies. 
However, we identify two major limitations in these designs. 
First, they lack scene awareness: memories from unrelated life scenes share the same retrieval space, which inflates the search space and introduces cross-scene interference.
Second, they encode each memory from a single perspective, making it difficult to retrieve complementary views of the same event.
In this paper, we propose the CreaMem architecture, which enables scene-aware memory organization by partitioning memory into several Life Scene Memories to reduce cross-scene interference at retrieval. To go beyond the single perspective and achieve cross-memory synergy, entries are dual-coded from both episodic and trait-based perspectives within each memory. We further devise a per-memory balanced sampling strategy at retrieval time. 
Extensive experiments on two long-term memory benchmarks show that CreaMem improves QA accuracy across all evaluation metrics, with particularly large gains on multi-hop reasoning performance, validating scene-aware partitioning and cross-memory synergy. To enhance reproducibility, we release our code in a public GitHub \href{https://github.com/Jacob0618/CreaMem}{repository}.

\end{abstract}

\section{Introduction}
Large language models (LLMs) are evolving from single-session tools into long-term personalized agents that accompany users across extended interactions~\cite{li2024personal}.
A core capability of such agents is long-term memory, which enables them to retain, organize, and retrieve information from prior interactions, thereby maintaining coherence across sessions~\cite{wang2024survey,guo2024large}. 
However, retention alone is insufficient. 
As interactions accumulate over weeks and months, the memory structure determines how the information is organized and whether the right information can be retrieved. 
Consequently, the architectural design of memory organization and retrieval plays a significant role in sustaining coherent long-term behavior across extended interactions~\cite{packer2023memgpt,park2023generative}, thus emerging as a central design problem for modern agent systems.

Existing memory systems address this design problem through various organization principles, which can be broadly categorized into three types: (1) Flat-storage methods~\cite{song2020mpnet,izacard2021contriever,lee2023mpc} retrieve from an unstructured pool; (2) Structure-imposing methods~(\citealp{pan2025secom}; \citealp{sarthi2024raptor}; \citealp{wang2025recursum}; \citealp{gutierrez2025hipporag2}; \citealp{xu2025amem}; \citealp{memgas2025}) organize memory through explicit structures such as summary trees or entity graphs; and (3) Memory-partitioning methods~\cite{li2026bmam,kang2025memoryos} decompose memory into several specialized components along axes like cognitive function or temporal scale. For example, MIRIX~\cite{wang2025mirix} separates memory into episodic, semantic, and procedural memories.

Although these methods differ in their organization principles, they overlook two aspects: none organizes memory by the user's life scenes, and few encode each experience along both an episodic and a trait-based perspective. These two gaps echo two well-established findings in cognitive psychology: (1) autobiographical memory is organized around lifetime contexts rather than abstract categories~\cite{conway2000construction,conway2005memory}; (2) episodic and semantic memory encode the same experience in complementary forms that cooperate during recall~\cite{tulving1972episodic,tulving2002episodic}.

We adopt both as design principles: memory is partitioned along the user's life scenes, and each experience is encoded twice. It is stored as a timeline entry in the Episodic Memory and as a trait entry in the relevant scene memory. For example, when a user mentions a weekend hike with their family, the system stores an episodic entry such as "the user went hiking on May 7" with surrounding event details, alongside a Life Scene trait entry such as "the user spends weekends on family outdoor activities." Traits are kept within scenes because user preferences are context-dependent: the same kind of event in a work setting would yield a different trait, and pooling them together would wash out these scene-specific patterns. At retrieval, a temporal query such as "when did the user go hiking?" is answered by the Episodic Memory entry alone, while a broader query such as "is the user likely to enjoy a camping trip?" draws from both memories, returning the event and its scene-specific trait as mutually complementary perspectives.

In this paper, we instantiate these principles as CreaMem, a scene-aware memory architecture for long-term personal agents. At storage time, a Meta Memory Manager routes each incoming message to the relevant scene memory and adds a paired entry to the Episodic Memory. At retrieval time, queries are issued to all memories, and a per-memory balanced sampling strategy ensures that both the timeline view and the scene-specific view contribute to the final returned context for response generation.

In summary, our contributions are threefold.
\begin{itemize}[leftmargin=*,nosep]
\item We partition memory along the user's life scenes, i.e., Life, Work, and Interest, together with a dedicated Episodic Memory, thereby reducing cross-scene interference at retrieval.
\item We encode each experience as both an episodic entry in the Episodic Memory and a trait entry in the relevant scene memory, and combine them at retrieval through per-memory balanced sampling.
\item Experiments on two long-term memory benchmarks show that CreaMem improves overall QA accuracy and multi-hop reasoning performance; ablation study further indicates that cross-memory synergy, not any single component, accounts for the majority of the observed gain.
\end{itemize}

\section{Related Work}

CreaMem draws from two bodies of prior work: agent memory systems for LLMs, and cognitive science theories of human memory organization.

\subsection{Memory for LLM Agents}
\label{sec:related_work}

LLMs face fundamental challenges in handling complex 
scenarios that require long-term coherence, where fixed-length contexts struggle to maintain continuity across dialogues with temporal gaps~\cite{wang2024survey,guo2024large,wu2026memoryera}. 
Full-context approaches~\cite{chen2023extending,brown2020gpt3,
achiam2023gpt4,touvron2023llama} place the dialogue history directly in the LLM's context window, but scale poorly as histories grow beyond the window~\cite{gao2024retrieval, liu2024lost,paulsen2025context}.

External memory systems address this by extracting and 
retrieving relevant content on demand. Flat-storage methods such as MemGPT~\cite{packer2023memgpt}, MemoryBank~\cite{zhong2023memorybank}, MPNet~\cite{song2020mpnet}, Contriever~\cite{izacard2021contriever}, and MPC~\cite{lee2023mpc} index memory entries in an undifferentiated pool for RAG-style retrieval~\cite{lewis2020rag}, where unrelated entries compete for retrieval slots. Structure-imposing methods 
mitigate this by adding summary hierarchies~\cite{wang2025recursum,sarthi2024raptor}, topical segments~\cite{pan2025secom}, knowledge 
graphs~\cite{gutierrez2025hipporag2,rasmussen2025zep,
pan2024unifying,hamilton2017graphsage}, or note-based 
links~\cite{xu2025amem,chhikara2025mem0}, yet remain confined to a single memory pool where unrelated contexts still compete.

Most closely related to our work, memory-partitioning 
architectures~\cite{wang2025mirix,kang2025memoryos,li2026bmam,m2a2026,evermemos2026,memma2026,zhou2025m2pa,lei2025robomemory,tiwari2026multilayered,yang2026plugmem,yu2025memagent} 
decompose memory into specialized components along 
axes such as temporal scale or cognitive function. 
For example, MemoryOS~\cite{kang2025memoryos} organizes memory into short-, mid-, and long-term levels, while MIRIX~\cite{wang2025mirix} separates episodic, semantic, and procedural memory.
This paradigm leaves two aspects unaddressed: memories 
from different life scenes still share the same space at retrieval, and each experience is encoded from a single perspective without a mechanism for combining complementary views.

\begin{figure*}[!t]
\centering
\includegraphics[width=\textwidth]{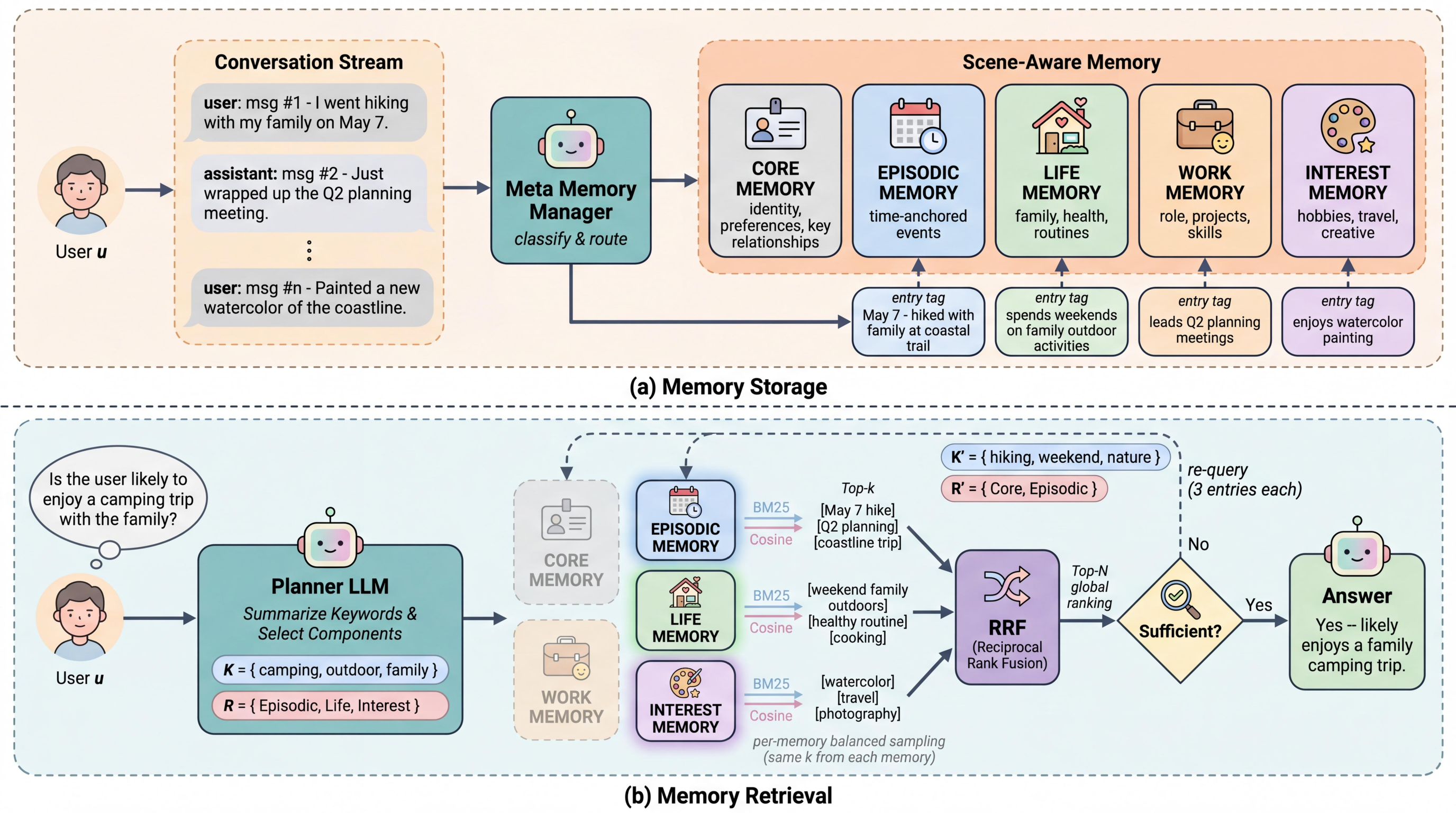}
\caption{Overview of the CreaMem architecture, consisting of a Meta Memory Manager, an Episodic Memory, three Life Scene Memories, and a Core Memory.}
\label{fig:architecture}
\end{figure*}

\subsection{Cognitive Foundations}
Two complementary lines of cognitive psychology inform CreaMem's design. Conway's Self-Memory System~\cite{conway2000construction,
conway2005memory} observes that autobiographical memories are organized by lifetime contexts rather than abstract categories, while Tulving's multiple memory systems~\cite{tulving1972episodic,tulving2002episodic} further observes that episodic and semantic memory encode the same experience from complementary perspectives and cooperate during retrieval. 

Inspired by these observations, CreaMem partitions memory along user life scenes and stores each experience as a timeline entry in the Episodic Memory and a trait entry in the relevant scene memory.


\section{CreaMem}


CreaMem is a scene-aware memory architecture for long-term personal agents that organizes user interactions along user's life scenes and combines complementary timeline with trait perspectives at retrieval, supporting coherent and personalized responses across extended conversations.

\subsection{Overview Architecture}
The overall architecture of CreaMem is illustrated in Figure~\ref{fig:architecture}. It consists of three modules: memory components, storage and retrieval.

\textbf{Memory Components:} This module defines the structural organization of memory in CreaMem. Memory comprises three components: an Episodic Memory that stores timeline entries of past events, a set of Life Scene Memories, Life, Work, and Interest, that store trait entries reflecting the user's behavior and preferences within each life scene, and a Core Memory that maintains a holistic user profile spanning all scenes.

\textbf{Memory Storage:} This module handles how each incoming message is written into memory. A Meta Memory Manager routes each message to the relevant Life Scene Memory and simultaneously adds a paired timeline entry to the Episodic Memory, recording each experience along both temporal and trait perspectives.

\textbf{Memory Retrieval:} This module retrieves relevant entries for a given query. Queries are issued to all selected memory components, and a per-memory balanced sampling strategy ensures that both the timeline view and the scene-specific view contribute to the returned context.





\subsection{Memory Components}
\label{sec:components}

\paragraph{Meta Memory Manager.}
The Meta Memory Manager serves as the central router of CreaMem. Upon receiving an incoming message, it classifies the content into one or more relevant scenes and dispatches the segment to the corresponding memories. Because a single message often carries information across multiple life scenes, e.g. ``I went to the gym with my colleague after work'', the Manager may route the same message to Life, Work, and Interest Memories simultaneously.

\paragraph{Planner LLM.} Given a user question, the Planner LLM selects memory components to query, outputs a set of keywords to drive retrieval, and generates the final response.

\paragraph{Core Memory.}
Core Memory maintains a stable user profile across scenes, storing identity attributes, preferences, key relationships, and other static reference information. It is injected into the prompt in full when selected by the Planner LLM.

\paragraph{Episodic Memory.}
Episodic Memory stores time-anchored events together with their surrounding narrative details. Each entry records a specific occurrence, such as a hike on 7 May, a promotion announcement, or concert attendance, along with the contextual details from which it was drawn. This component supports queries that require concrete grounding in particular past moments.

\paragraph{Life Scene Memories.}
The three Life Scene Memories capture trait-level knowledge about the user within a distinct life scene. \textbf{Life Memory} covers personal-life aspects such as family, health, daily routines, and food. \textbf{Work Memory} covers professional and academic aspects such as the user's role, ongoing projects, skills, and career goals. \textbf{Interest Memory} covers leisure aspects such as hobbies, entertainment, travel, and creative pursuits. Each entry within a Life Scene Memory is a trait: a distilled, scene-conditional inference about the user, abstracted from one or more underlying messages in the dialogue history.

\subsection{Memory Storage}
\label{sec:storage}

Each incoming message $m$ is processed through a three-step pipeline. The Meta Memory Manager first classifies $m$ into a subset of relevant memory components $R(m) \subseteq \{\text{Core}, \text{Episodic}, \text{Life}, \text{Work}, \text{Interest}\}$. For each selected memory component $r \in R(m)$, a corresponding LLM-based extractor $f_{\text{LLM}}^{r}$ produces candidate memory entries. For each candidate, the system retrieves the most similar existing entries via embedding-based lookup, and the LLM determines whether to merge the candidate into an existing entry or insert it as a new entry into the corresponding SQLite table.

\paragraph{Routing policy.}
The Manager follows a \emph{wide-entry, strict-filtering} policy: when scene classification is uncertain, multiple memory components are triggered to prevent silent information loss. A single message may therefore be routed to several memory components at once.

\paragraph{Core Memory.}
Core Memory stores the user's profile as a free-text block. For each routed message, the LLM appends an extracted update $v = f_{\text{LLM}}^{\text{core}}(m)$ to the block. When the block reaches 90\% of its character limit, the LLM automatically compresses it to free space for further updates.
\paragraph{Episodic Memory.} Each Episodic entry is a 
structured tuple
\begin{equation}
M_{\text{epi}}^{(i)} = \big(\, \mathrm{eid}_i,\; t_i,\; a_i,\; e_i,\; s_i,\; d_i,\; \mathbf{v}_{s,i},\; \mathbf{v}_{d,i} \,\big),
\end{equation}
where $\mathrm{eid}_i$ is a unique episodic entry identifier, $t_i$ is the event timestamp, $a_i \in \{\text{user}, \text{assistant}\}$ the actor, $e_i$ the event type, $s_i$ a one-sentence summary, $d_i$ the full event context, and $\mathbf{v}_{s,i} = f_{\text{emb}}(s_i)$, $\mathbf{v}_{d,i} = f_{\text{emb}}(d_i)$ are dense embeddings of the summary and details produced by an embedding function $f_{\text{emb}}$. A single message may yield multiple entries:
\begin{equation}
\{M_{\text{epi}}^{(i)}\}_{i=1}^{n} = f_{\text{LLM}}^{\text{epi}}(m).
\end{equation}

\paragraph{Life Scene Memories.}
The three Life Scene Memories share a unified entry schema
\begin{equation}
M_{\text{scene}}^{(i)} = \big(\, \mathrm{sid}_i,\; c_i,\; w_i,\; \mathbf{v}_i \,\big), \quad \mathbf{v}_i = f_{\text{emb}}(c_i),
\end{equation}
where $\mathrm{sid}_i$ is a unique entry identifier within its life scene memory, $\text{scene} \in \{\text{Life}, \text{Work}, \text{Interest}\}$ indexes the three memories, $c_i$ is an extracted trait content, $w_i \in [0, 1]$ its importance score assigned by the LLM, and $\mathbf{v}_i$ the embedding of the content.

\subsection{Memory Retrieval}
\label{sec:retrieval}

Given a user question $q$, CreaMem performs three operations: memory selection by a planning LLM, per-memory hybrid retrieval with balanced sampling, and global rank fusion.

\paragraph{Memory selection.}
The Planner LLM examines the question and selects a subset of relevant memories $R \subseteq \{\text{Core}, \text{Episodic}, \text{Life}, \text{Work}, \text{Interest}\}$ together with a keyword set $\mathcal{K}$ of 3 to 6 keywords used for retrieval:
\begin{equation}
(\mathcal{K},\, R) = f_{\text{LLM}}^{\text{plan}}(q).
\end{equation}

\paragraph{Per-memory balanced retrieval.}
For each selected memory $r \in R$, two parallel retrieval paths are applied: BM25 over textual content and cosine similarity over the query embedding $\mathbf{v}_q = f_{\text{emb}}(\mathcal{K})$. Each path returns the top-$k$ candidates from $r$, denoted $C_r^{\text{bm}}$ and $C_r^{\text{emb}}$ respectively. Because every memory contributes the same number of candidates, no single memory dominates the pool. The combined candidate set $C$ is formed by taking the union of all per-memory candidates and deduplicating by memory ID and content prefix:
\begin{equation}
C = \mathrm{dedup}\bigg(\bigcup_{r \in R} \big(C_r^{\text{bm}} \cup C_r^{\text{emb}}\big)\bigg).
\end{equation}

\paragraph{Reciprocal Rank Fusion.}
Candidates in $C$ are re-ranked globally by both signals and fused via Reciprocal Rank Fusion~\cite{cormack2009rrf}. For each $c \in C$,
\begin{equation}
\mathrm{RRF}(c) = \frac{1}{k_0 + \mathrm{rank}_{\text{bm}}(c)} + \frac{1}{k_0 + \mathrm{rank}_{\text{emb}}(c)},
\end{equation}
where $k_0$ is a smoothing constant. The top-$N$ candidates by RRF score are inserted into the prompt as retrieved context for response generation. If the Planner LLM judges the context insufficient, it selects memories to re-query at 3 entries each.




\begin{table*}[t]
\centering
\small
\begin{tabular}{l|c|c|c|c|c|c|c|c}
\toprule
Model & 4o-J & F1 & B-4 & R-1 & R-2 & R-L & BS & Avg.\,Tokens \\
\midrule
\multicolumn{9}{c}{\textit{LoCoMo}} \\
\midrule
Full History & 33.43 & 12.23 & 1.84 & 12.70 & 5.66 & 11.73 & 84.07 & 20{,}078  \\
MPNet~\cite{song2020mpnet} & 38.07 & 14.44 & 2.35 & 14.90 & 6.83 & 13.90 & 84.42 & 2{,}472  \\
Contriever~\cite{izacard2021contriever} & 40.33 & 15.66 & 2.67 & 16.01 & 7.68 & 15.00 & 84.65 & 2{,}348  \\
MPC~\cite{lee2023mpc} & 40.38 & 14.81 & 1.99 & 15.10 & 6.83 & 14.13 & 84.42 & 2{,}683  \\
RecurSum~\cite{wang2025recursum} & 22.56 & 9.14 & 0.99 & 9.82 & 3.38 & 8.98 & 83.45 & 3{,}074  \\
SeCom~\cite{pan2025secom} & 44.21 & 13.79 & 2.30 & 14.28 & 6.17 & 13.30 & 84.04 & \textbf{1{,}021}  \\
HippoRAG~2~\cite{gutierrez2025hipporag2} & 45.62 & 16.66 & 2.91 & 17.01 & 8.27 & 15.93 & 84.88 & 2{,}991  \\
RAPTOR~\cite{sarthi2024raptor} & 31.72 & 14.55 & 2.88 & 15.09 & 7.49 & 14.18 & 84.48 & 1{,}931  \\
A-Mem~\cite{xu2025amem} & 40.81 & 14.72 & 2.83 & 16.22 & 7.71 & 14.89 & 84.72 & 3{,}042  \\
MemGAS~\cite{memgas2025} & 41.07 & 17.66 & 3.61 & 18.00 & 8.93 & 16.99 & 85.13 & 2{,}825  \\
MemoryOS~\cite{kang2025memoryos} & 43.96 & 16.92 & 3.59 & 17.49 & 7.69 & 16.12 & 84.84 & 2{,}833 \\
\midrule
\textbf{CreaMem (Ours)} & \textbf{54.61} & \textbf{19.34} & \textbf{4.21} & \textbf{20.01} & \textbf{9.04} & \textbf{18.50} & \textbf{85.30} & 2{,}805  \\
\midrule
\multicolumn{9}{c}{\textit{LongMemEval-S}} \\
\midrule
Full History & 50.60 & 11.48 & 1.40 & 12.10 & 5.47 & 10.85 & 83.07 & 103{,}137 \\
MPNet~\cite{song2020mpnet} & 53.20 & 13.96 & 2.21 & 14.49 & 6.78 & 12.93 & 83.72 & 8{,}173 \\
Contriever~\cite{izacard2021contriever} & 55.40 & 13.78 & 2.21 & 14.46 & 6.93 & 12.89 & 83.70 & 8{,}286 \\
MPC~\cite{lee2023mpc} & 53.80 & 13.60 & 1.74 & 14.27 & 6.49 & 12.95 & 83.49 & 8{,}457 \\
RecurSum~\cite{wang2025recursum} & 35.40 & 12.29 & 2.09 & 13.01 & 5.55 & 11.52 & 83.60 & 8{,}853 \\
SeCom~\cite{pan2025secom} & 56.00 & 12.95 & 2.25 & 13.80 & 6.09 & 11.93 & 83.51 & \textbf{2{,}741} \\
HippoRAG~2~\cite{gutierrez2025hipporag2} & 57.60 & 14.73 & 2.15 & 15.30 & 7.36 & 13.83 & 83.86 & 8{,}530 \\
RAPTOR~\cite{sarthi2024raptor} & 32.20 & 12.08 & 1.90 & 12.73 & 5.82 & 11.25 & 83.50 & 6{,}254 \\
A-Mem~\cite{xu2025amem} & 55.60 & 13.73 & 2.11 & 14.82 & 6.81 & 12.98 & 83.88 & 9{,}018 \\
MemGAS~\cite{memgas2025} & 60.20 & 20.38 & 4.22 & 21.05 & 10.47 & 19.47 & 85.21 & 8{,}829 \\
\midrule
\textbf{CreaMem (Ours)} & \textbf{66.40} & \textbf{20.71} & \textbf{4.79} & \textbf{21.82} & \textbf{11.13} & \textbf{19.78} & \textbf{85.36} & 7{,}250 \\
\bottomrule
\end{tabular}
\caption{Main results on LoCoMo (top) and LongMemEval-S (bottom). Best per column in bold.}
\label{tab:main_results}
\end{table*}

\section{Experiments}
\subsection{Evaluation Setup}
\label{sec:eval-methodology}

\paragraph{Datasets.}
We evaluate CreaMem on LoCoMo~\cite{maharana2024locomo} and LongMemEval-S~\cite{wu2025longmemeval}. LoCoMo targets long-term conversational memory, with 10 ultra-long dialogues of around 300 turns and 9K tokens each, and 1{,}540 questions across four types: Single-hop, Multi-hop, Open-domain, and Temporal. LongMemEval-S contains 500 questions testing memory retention across multi-session dialogues.

\paragraph{Evaluation Metrics.}
We follow the MemGAS evaluation pipeline~\cite{memgas2025} for consistency with prior work. Following standard practice, our main metric is 4o-Judge accuracy, where GPT-4o serves as an LLM judge to assess response correctness against ground-truth answers. We additionally report Token F1~\cite{rajpurkar2016squad}, BLEU-4, ROUGE-1/2/L, and BertScore for comprehensive lexical and semantic comparison.

\paragraph{Compared Methods.}
We compare CreaMem with representative methods covering the major paradigms of long-term conversational memory. To isolate the effect of memory architecture, all methods use GPT-4o-mini as their backbone LLM under an identical evaluation protocol.

\textbf{Full History}: This baseline places the entire dialogue history directly into the LLM context window without any retrieval or memory organization, serving as an unstructured upper reference.

\textbf{MPNet}~\cite{song2020mpnet}: A pretrained sentence encoder used as a dense retriever, indexing memory entries in a single, undifferentiated vector pool retrieved via cosine similarity.

\textbf{Contriever}~\cite{izacard2021contriever}: An unsupervised dense retriever that produces general-purpose embeddings for similarity-based retrieval over an undifferentiated memory pool.

\textbf{MPC}~\cite{lee2023mpc}: A prompted memory system that organizes conversational history through LLM-driven extraction without imposing explicit hierarchical or graph structure.

\textbf{RecurSum}~\cite{wang2025recursum}: Recursively summarizes past dialogue segments into a hierarchical summary structure for long-term retention.

\textbf{SeCom}~\cite{pan2025secom}: Partitions dialogue into topical segments and retrieves at the segment granularity for coherent context recall.

\textbf{RAPTOR}~\cite{sarthi2024raptor}: Builds a tree of recursive abstractive summaries that enables retrieval at multiple abstraction levels.

\textbf{HippoRAG 2}~\cite{gutierrez2025hipporag2}: Constructs a knowledge graph over entities and relations extracted from past dialogues, and supports relational retrieval over the graph inspired by the hippocampal memory model.

\textbf{A-Mem}~\cite{xu2025amem}: An agentic memory system that maintains Zettelkasten-style interconnected notes with self-reflective links, enabling continuous memory evolution.

\textbf{MemGAS}~\cite{memgas2025}: A multi-granularity memory system that maintains associations across abstraction levels for fine-grained retrieval.

\textbf{MemoryOS}~\cite{kang2025memoryos}: An OS-inspired memory system that partitions conversational memory along temporal scale into short-, mid-, and long-term tiers with heat-based promotion.

Notably, the chosen baselines are representative methods from each of the three paradigms identified in 
Section~\ref{sec:related_work}. Results for MemoryOS on 
LongMemEval-S are unavailable due to high runtime, as 
its memory organization requires a large number of LLM 
calls per turn, which we found impractical to reproduce 
on LongMemEval-S.

\begin{figure}[!t]
\centering
\includegraphics[width=\columnwidth]{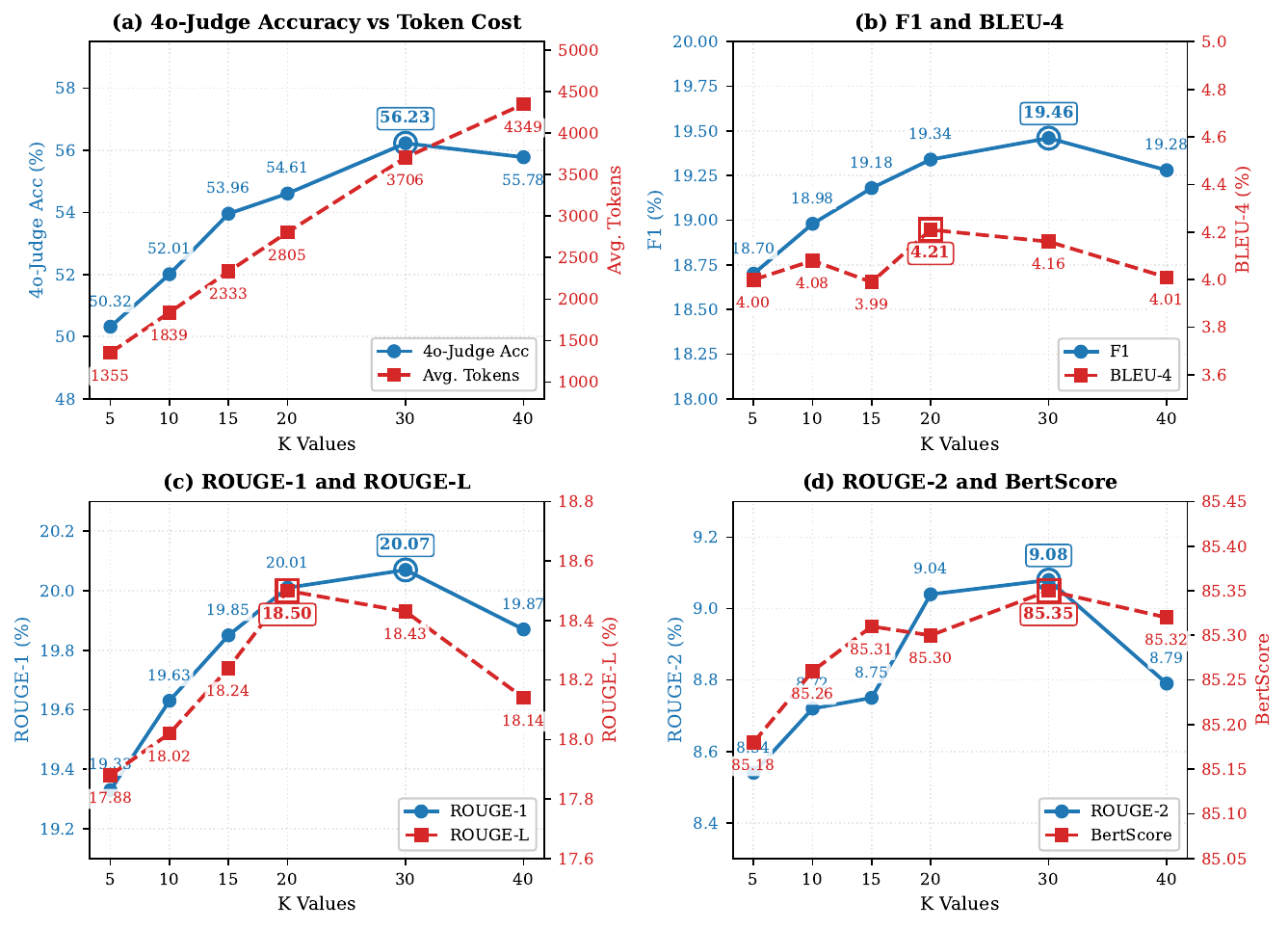}
\caption{Sensitivity of retrieval budget $k$ on LoCoMo. (a) 4o-Judge accuracy and token cost. (b) F1, BLEU-4. (c) ROUGE-1/L. (d) ROUGE-2, BertScore.}
\label{fig:k_analysis}
\end{figure}

\paragraph{Implementation Details.}
We implement CreaMem with GPT-4o-mini as the backbone LLM for the Meta Memory Manager, all memory-specific extractors, and the LLM Planner, with temperature set to 0 throughout. Memory is persisted in a local SQLite database, and text embeddings are produced by \texttt{text-embedding-3-small}. Batch extraction triggers once 3 or more user messages accumulate.

At retrieval, the Planner LLM outputs 3 to 6 keywords per query; per-memory top-k and final top-N are 20 in LoCoMo and 40 in LongMemEval-S. Figure~\ref{fig:k_analysis} shows that k=30 yields marginally higher accuracy in LoCoMo at the cost of substantially more retrieved tokens. We deliberately adopt the lower k=20 to match the token budget of the baselines in Table~\ref{tab:main_results}, ensuring that the comparison isolates memory organization from context size. The RRF smoothing constant $k_0$ is 60 following standard practice~\cite{cormack2009rrf}. Core Memory has a character limit of $L = 2000$ with automatic compression at 90\% capacity.

For evaluation, GPT-4o serves as the LLM judge for the 4o-Judge metric. We adopt the QA and judge prompts of MemGAS~\cite{memgas2025} for fair comparison with prior work. Full prompt templates are in Appendix~\ref{app:prompts}.

\begin{figure*}[!t]
\centering
\includegraphics[width=\textwidth]{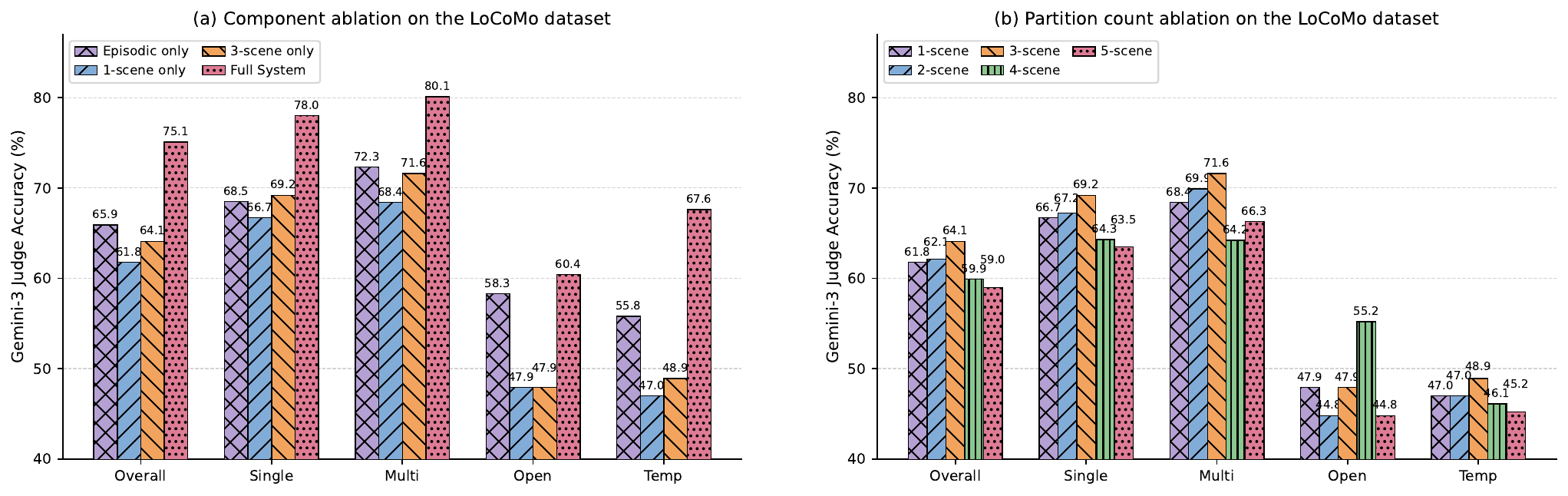}
\caption{Ablation results on the LoCoMo dataset. (a) Component ablation. (b) Partition count ablation.}
\label{fig:ablation}
\end{figure*}

\begin{figure*}[!t]
\centering
\includegraphics[width=\textwidth]{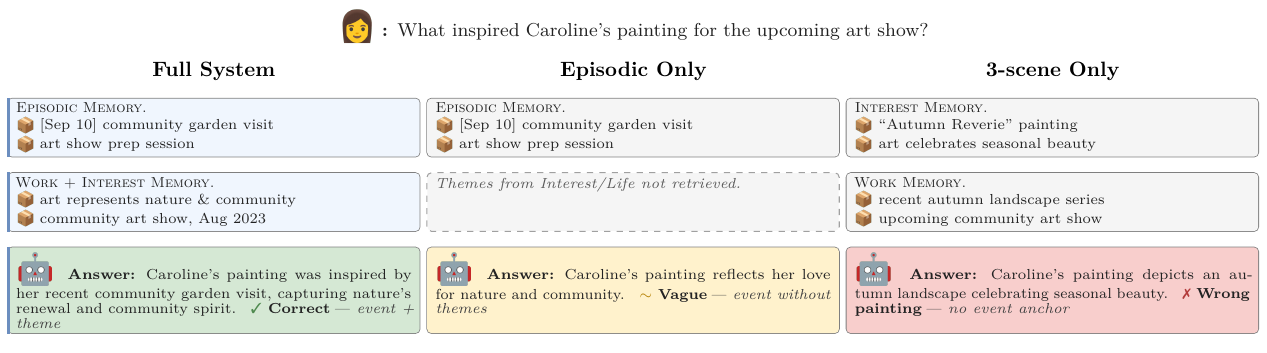}
\caption{Case study on a multi-hop LoCoMo question. The Full System integrates episodic and scene-level evidence, while either ablation produces incomplete or temporally ungrounded answers.}
\label{fig:case_study}
\end{figure*}

\subsection{Main Results}
Table~\ref{tab:main_results} presents the aggregate results on LoCoMo and LongMemEval-S. We have the following observations:

(1) \textbf{ Flat-storage methods lack organizational 
structure for retrieval.} MPNet, Contriever, and MPC index all memory entries in a single undifferentiated pool. Their accuracy forms a stable middle tier on both benchmarks, indicating that pure similarity-based retrieval has a clear ceiling without further organization.

(2) \textbf{Structure-imposing methods narrow 
but do not close the gap.} HippoRAG~2 and MemGAS lead 
this group on LoCoMo and LongMemEval-S respectively, 
while summary-based variants such as RecurSum and RAPTOR trade factual specificity for compactness and form the lowest tier. The shared ceiling across this group reflects a common limitation: structure is imposed within a single, undifferentiated memory pool, so entries from unrelated life scenes still compete for the same retrieval space, producing cross-scene interference.

(3) \textbf{Memory-partitioning by temporal scale 
does not relieve cross-scene interference.} MemoryOS 
partitions memory along temporal scale rather than along user's life scenes, and on LoCoMo lands in the same performance band as the strongest structure-imposing baselines. However, entries from unrelated life scenes can still co-occur within the same tier, leaving cross-scene interference unresolved at retrieval time.

(4) \textbf{CreaMem achieves the best performance across both datasets and all metrics.} It surpasses HippoRAG~2 and MemGAS, the strongest baselines on LoCoMo and LongMemEval-S respectively, and the lead is consistent across F1, BLEU-4, ROUGE-1/2/L, and BertScore rather than concentrated on a single metric. This breadth indicates that the gain comes from selecting more relevant content, not from surface-level lexical overlap.

(5) \textbf{The gains come from organization, not from a larger context budget.} CreaMem's token consumption is on par with the strongest structured baselines on LoCoMo and below most baselines on LongMemEval-S, where it uses an order of magnitude fewer tokens than Full History, yet it delivers the highest accuracy. Improvement therefore arises from how memory is organized and retrieved, not from feeding the LLM more raw context.

\subsection{Ablation Study}
We conduct three ablations on the full LoCoMo corpus using Gemini-3 as both QA backbone and judge. Since architectural benefits on an early-generation backbone like GPT-4o-mini do not automatically transfer to newer LLMs with stronger reasoning and context handling, we re-run all configurations on Gemini-3 to verify that CreaMem's gains persist on a substantially newer backbone. Across all configurations, the retrieval budget is fixed at 20 memory entries per query and the underlying memory entries are identical: ablated variants reorganize the same extracted content rather than re-extracting. Core Memory and the re-query step are held out across all variants, as both are orthogonal to the scene-partitioning axis under study. Observed differences therefore reflect how memory is structured at retrieval, not how much information is stored or surfaced into context.

\textbf{Component Ablation.} Figure~\ref{fig:ablation}(a) compares four configurations testing whether CreaMem's two design choices, scene partitioning and dual encoding, are necessary or redundant.

First, we test whether three scene memories add value beyond a single undifferentiated pool, since partitioning could simply complicate retrieval. We compare \textbf{3-scene only}, which retains the three scene memories without Episodic Memory, with \textbf{1-scene only}, which merges them into a single pool while keeping every trait entry verbatim. Under identical content and retrieval budget, 3-scene outperforms 1-scene by 2.3 points, showing that scene-based organization is not redundant against flat storage at retrieval time.

Second, we test whether storing each experience as both an episodic event and a trait merely duplicates information. We compare \textbf{Full System}, which couples Episodic Memory with three scene memories, against \textbf{Episodic only}, which retains timeline and event details alone. Full System reaches 75.1\% overall, 9.2 points above Episodic only, with the gap widening to over 11 points on Temporal questions. This gain is not from feeding more context: at the same retrieval budget, Episodic only consumes 42\% more tokens than Full System, since each episodic entry's details field preserves the full event narrative and surrounding context, carrying even more information per entry than Full System. Dual encoding therefore improves accuracy while reducing token consumption. Figure~\ref{fig:case_study} illustrates this on a multi-hop case where ablating either side yields incomplete or temporally ungrounded answers.

Third, we test whether finer partitioning further improves performance. Figure~\ref{fig:ablation}(b) compares five partition counts under a fixed retrieval budget of 20 entries per query, evenly distributed across memories, matching the baseline cost in Table~\ref{tab:main_results}. \textbf{1-scene} merges all scenes, \textbf{2-scene} merges Life and Work, \textbf{3-scene} is our default, \textbf{4-scene} splits Life into ``family''/``other'', and \textbf{5-scene} further splits Interest into ``hobby''/``entertainment'', with splits produced by LLM re-tagging existing entries. Among the partition counts tested under this retrieval budget, the 3-scene Life/Work/Interest configuration achieves the best balance: coarser partitions force unrelated scene content to compete for the same retrieval slots, while finer partitions over-fragment memory and dilute per-memory information density.

\subsection{Hyperparameter Analysis}
We analyze the impact of the per-memory retrieval budget k on model performance. As shown in Figure~\ref{fig:k_analysis}, accuracy is non-monotone in k: 4o-Judge peaks at k=30 and declines at k=40, with the same pattern across F1, ROUGE-1/2/L, and BertScore, while token cost grows nearly linearly. Beyond a modest budget, additional entries act as noise rather than signal, degrading answers despite higher cost. 

\subsection{Architectural Validation}
\label{sec:validation}
CreaMem rests on the prerequisite that the Life/Work/Interest taxonomy covers the bulk of personal memory content. We validate this empirically by sampling 1{,}000 dialogue turns from the 10 LoCoMo conversations, 100 turns per conversation. Each turn is independently judged by GPT-4o and three human raters on whether it falls within one of the three scenes. As shown in Table~\ref{tab:coverage}, coverage is consistently high across judges, with strong LLM--human agreement. The remaining gap comes from turns that are hard to assign to any scene, such as greetings or small talk. Overall, most personal content in long-term dialogue fits into the three scenes validating CreaMem's design prerequisite.

\begin{table}[!t]
\centering
\small
\begin{tabular}{lc}
\toprule
Judge & Coverage (\%) \\
\midrule
GPT-4o            & 71.00 \\
Human Rater A     & 83.00 \\
Human Rater B     & 90.10 \\
Human Rater C     & 80.50 \\
\midrule
LLM--Human agreement & 77.20 \\
\bottomrule
\end{tabular}
\caption{Coverage of the Life/Work/Interest taxonomy over 1{,}000 sampled LoCoMo dialogue turns, 100 per conversation, judged by GPT-4o and three human raters.}
\label{tab:coverage}
\end{table}

\subsection{Failure Analysis}
\label{sec:failure-analysis}
Aggregate improvement does not imply per-question dominance: CreaMem still fails on some questions answered correctly by competing methods. In a Single-hop question about Nate's second tournament, CreaMem stored the event and the game name, \emph{Street Fighter}, in separate entries. Only the event entry appeared in the retrieved context, whereas SeCom retained the contiguous dialogue segment and answered correctly. In an Open-domain question about John's endorsement deal, CreaMem retrieved both the unnamed outdoor-gear deal and his earlier interest in Under Armour. Response generation nevertheless failed to connect the entries, whereas MemoryOS produced the reference answer. The first case reflects incomplete retrieval of linked entries; the second reflects failure to combine retrieved entries after retrieval. These cases motivate links to underlying dialogue turns, fallback to relevant dialogue segments, and explicit evidence combination. Appendix~\ref{app:failure-cases} summarizes the source evidence, model outputs, and one additional retrieval failure.

\section{Conclusion}
We presented CreaMem, a scene-aware memory architecture motivated by autobiographical memory organization and dual coding in cognitive psychology. It partitions memory along the user's life scenes alongside Episodic and Core Memory, dual-encoding each experience and combining views via per-memory balanced sampling. Experiments on LoCoMo and LongMemEval-S show consistent gains in accuracy and multi-hop reasoning at comparable token cost, suggesting that the key lever for coherent long-term agents is how memory is organized, not how much context is supplied.

\section*{Limitations}
CreaMem has several limitations. First, the fixed Life/Work/Interest taxonomy is a deliberately coarse organization rather than a universal or optimal ontology. Although our results support this partition on the evaluated benchmarks, it may not transfer uniformly across users, cultures, or specialized domains. Future work will investigate configurable or automatically induced scene partitions that adapt to individual users while retaining interpretable routing.

Second, CreaMem incurs additional system overhead. End-to-end latency averages 2.9 seconds per query, compared with 1.98 seconds for A-Mem. The sufficiency check can also trigger a second retrieval round, improving recall at the cost of additional latency and tokens. Query-adaptive pruning, asynchronous retrieval, and learned stopping criteria may reduce this overhead.

Several evaluation challenges remain for LLM-extraction memory systems. Write-time costs are difficult to compare because methods differ in batching granularity, module prompts, and the abstraction level of stored entries. Retrieval metrics such as Recall@$k$ and NDCG@$k$ also assume direct alignment between retrieved items and reference evidence. This assumption does not hold when memory entries summarize or combine multiple dialogue turns. Finally, extraction behavior depends on prompt and model choice. Standardized write-time accounting, source-aligned retrieval evaluation, and method-agnostic extraction protocols remain important directions for future work.

\section*{Acknowledgments}
This work is supported by the Early Career Scheme (No.CityU 21219323) and the General Research Fund (No.CityU 11220324) of the University Grants Committee (UGC), the NSFC Young Scientists Fund (No.9240127), and the Donation for Research Projects (No.9229216).


\bibliography{creaseed_refs}
\appendix
\section{Supplementary Analyses}
\label{app:failure-cases}
\label{app:supplementary-experiments}

\paragraph{Category-level and open-weight evaluation.}
Under the matched GPT-4o-mini/GPT-4o protocol, CreaMem leads Overall, Multi-hop, and Temporal accuracy, whereas SeCom and MemoryOS lead Single-hop and Open-domain, respectively (Table~\ref{tab:category-breakdown}a). Its margins over the strongest baseline are 2.83 points on Multi-hop and 19.93 points on Temporal questions. With Qwen3.6-35B-A3B, CreaMem leads by 5.97 points overall and remains strongest on Multi-hop, Open-domain, and Temporal questions (Table~\ref{tab:category-breakdown}b), supporting cross-backbone transfer but not uniform dominance.

\paragraph{Routing, extraction, and taxonomy reliability.}
Across 998 jointly judged LoCoMo turns, routing reaches 67.94\% exact match, 94.79\% partial match, and 74.53/73.65 Micro/Macro-F1 (Table~\ref{tab:diagnostic-audits}). At 40\% synthetic misrouting, QA declines only 1.23 points, from 54.61\% to 53.38\%. In a 500-memory audit, Gemini~3.1~Pro/GPT-4o find 90.40/93.60\% of entries retain a supported central fact, while 9.60/6.40\% contain an unsupported central claim. Among 107 vertical-domain LongMemEval-S turns, exact routing is 68.22\% and at-least-one-scene coverage is 99.03\%. Coverage is therefore broad, although routing and extraction errors remain.

\paragraph{Allocation and architectural ablations.}
Dynamic allocation reaches 55.26\% versus 55.00\% for balanced sampling ($+0.26$ points; 95\% CI $[-0.78,1.30]$) but adds 0.69 LLM calls per question. Under the matched GPT-4o-mini/GPT-4o setup, Full CreaMem reaches 53.24\%, versus 44.03\% for Episodic-only and 40.32--41.43\% for scene-only variants. RRF changes overall accuracy by only $+0.06$ and $-0.91$ points across two backbones, with both confidence intervals spanning zero. The main gain therefore arises from scene-aware dual encoding rather than allocation or fusion alone.

\paragraph{Representative failure cases.}
Table~\ref{tab:failure-cases} distinguishes failures to retrieve linked or implicit evidence from failures to combine evidence already present in the context.

\setcounter{dbltopnumber}{3}
\renewcommand{\dbltopfraction}{0.95}
\renewcommand{\dblfloatpagefraction}{0.70}
\makeatletter
\setlength{\@dblfptop}{0pt}
\setlength{\@dblfpsep}{11pt}
\setlength{\@dblfpbot}{0pt plus 1fil}
\makeatother

\begin{table*}[!t]
\centering
\footnotesize
\setlength{\tabcolsep}{4.5pt}
\renewcommand{\arraystretch}{0.98}
\textit{(a) GPT-4o-mini backbone and GPT-4o judge.}\\[-2pt]
\begin{tabular*}{\textwidth}{@{\extracolsep{\fill}}lccccc@{}}
\toprule
Method & Overall & Single-hop & Multi-hop & Open-domain & Temporal \\
\midrule
CreaMem     & \textbf{54.61} & 66.11 & \textbf{28.72} & 38.54 & \textbf{52.02} \\
HippoRAG~2  & 47.66 & 63.73 & 24.11 & 35.42 & 29.91 \\
SeCom       & 45.58 & \textbf{67.66} & 23.40 & 34.38 & 10.59 \\
MemoryOS    & 43.96 & 54.70 & 25.89 & \textbf{42.71} & 32.09 \\
MemGAS      & 43.38 & 57.19 & 23.05 & 33.33 & 28.04 \\
\bottomrule
\end{tabular*}

\vspace{5pt}
\textit{(b) Qwen3.6-35B-A3B backbone with thinking disabled.}\\[-2pt]
\begin{tabular*}{\textwidth}{@{\extracolsep{\fill}}lcccccc@{}}
\toprule
Method & Overall & Single-hop & Multi-hop & Open-domain & Temporal & Avg.\,Tokens \\
\midrule
CreaMem     & \textbf{57.66} & 66.11 & \textbf{31.21} & \textbf{39.58} & \textbf{64.17} & 2{,}915.64 \\
HippoRAG~2  & 51.69 & 64.21 & 29.79 & 29.17 & 44.86 & 2{,}482.88 \\
SeCom       & 47.01 & \textbf{69.56} & 26.60 & 35.42 & 9.35 & 907.19 \\
MemoryOS    & 39.35 & 50.42 & 19.86 & 36.46 & 28.35 & 3{,}087.18 \\
MemGAS      & 37.79 & 46.49 & 18.79 & 35.41 & 32.40 & 3{,}214.00 \\
\bottomrule
\end{tabular*}
\caption{Category-level LoCoMo results under closed and open-weight backbones. Values are 4o-Judge accuracy (\%); average QA tokens are reported for the Qwen evaluation. Best results within each backbone are bold.}
\label{tab:category-breakdown}
\label{tab:qwen-open-weight}
\end{table*}

\begin{table*}[!t]
\centering
\footnotesize
\setlength{\tabcolsep}{5pt}
\renewcommand{\arraystretch}{1.00}
\begin{tabular}{@{}p{0.19\textwidth}p{0.21\textwidth}p{0.52\textwidth}@{}}
\toprule
\raggedright Analysis & \raggedright Protocol & Result \\
\midrule
\raggedright\textbf{Routing quality} & \raggedright 998 jointly judged LoCoMo turns & Exact 67.94\%; partial 94.79\%; Micro/Macro-F1 74.53/73.65\%. \\
\addlinespace[2pt]
\raggedright\textbf{Synthetic misrouting} & \raggedright 1{,}540 questions; Work $\rightarrow$ Interest & Accuracy at 0/10/20/40\% corruption: 54.61/54.61/54.29/53.38\%. \\
\addlinespace[2pt]
\raggedright\textbf{Extraction faithfulness} & \raggedright 500 memories; Gemini~3.1~Pro/GPT-4o judges & Supported central fact: 90.40/93.60\%; fully supported: 71.20/73.40\%; unsupported central claim: 9.60/6.40\%. \\
\addlinespace[2pt]
\raggedright\textbf{Scene taxonomy} & \raggedright 107 vertical-domain LongMemEval-S turns & Conservative judge intersection: 68.22\% exact routing and 99.03\% at-least-one-scene coverage among non-empty references. \\
\addlinespace[2pt]
\raggedright\textbf{Retrieval allocation} & \raggedright 1{,}540 paired LoCoMo questions & Balanced 55.00\% versus Dynamic 55.26\%; $\Delta=+0.26$ pp, 95\% CI $[-0.78,1.30]$; Dynamic adds 0.69 LLM calls/question. \\
\addlinespace[2pt]
\raggedright\textbf{Components} & \raggedright GPT-4o-mini/GPT-4o; no Core Memory & Full System 53.24\%; Episodic-only 44.03\%; scene-only variants 40.32--41.43\%. \\
\addlinespace[2pt]
\raggedright\textbf{RRF vs. BM25-only} & \raggedright Gemini-3 and GPT-4o-mini/GPT-4o & $+0.06$ pp, 95\% CI $[-2.01,2.08]$; and $-0.91$ pp, 95\% CI $[-2.99,1.10]$, respectively. \\
\bottomrule
\end{tabular}
\caption{Reliability audits and supplementary ablations. Routing, misrouting, allocation, component, and RRF results use LoCoMo; the scene-taxonomy audit uses LongMemEval-S. Values follow the rebuttal protocols.}
\label{tab:diagnostic-audits}
\label{tab:supplementary-ablations}
\end{table*}

\begin{table*}[!t]
\centering
\footnotesize
\setlength{\tabcolsep}{4.5pt}
\renewcommand{\arraystretch}{1.02}
\begin{tabular}{@{}p{0.25\textwidth}p{0.37\textwidth}p{0.30\textwidth}@{}}
\toprule
Question (reference) & CreaMem behavior & Correct baseline and diagnosis \\
\midrule
\textbf{Single-hop:} Nate's second tournament (\emph{Street Fighter}) & The event and game were stored separately, but only the event entry was retrieved; the response stated that the game was unknown. & SeCom retained D10:4--D10:6 and answered correctly. Linked entries were not retrieved together. \\
\addlinespace
\textbf{Open-domain:} John's outdoor-gear endorsement (\emph{Under Armour}) & Both the unnamed deal and John's earlier interest in Under Armour were retrieved, but the response did not connect them. & MemoryOS answered correctly. Retrieved evidence was not combined during response generation. \\
\addlinespace
\textbf{Temporal:} Holiday season of Evan's wedding (\emph{Christmas}) & The date-bearing wedding session was absent from the context, so the response stated that the holiday could not be determined. & MemoryOS answered correctly. Implicit temporal evidence was not retrieved. \\
\bottomrule
\end{tabular}
\caption{Representative LoCoMo questions answered incorrectly by CreaMem but correctly by a baseline. The cases illustrate distinct failure stages rather than their prevalence.}
\label{tab:failure-cases}
\end{table*}

\raggedbottom
\section{Prompt Templates}
\label{app:prompts}

For brevity, we list only the key prompts of CreaMem below; additional prompts are documented in the released code. Prompts A.1--A.4 cover CreaMem's routing and extraction logic and are abridged for readability. Prompts A.5 and A.6 reproduce the QA and judge prompts from MemGAS~\cite{memgas2025} for completeness.

\bigskip

\begin{promptbox}{Meta Memory Manager (Storage Routing)}
\textbf{Role.} You are the Meta Memory Manager of CreaMem, responsible for routing batched user messages to the appropriate memory components.

\textbf{Task.} Analyze a batch of user messages and call \texttt{trigger\_memory\_update(memory\_types=[...])} with one or more types from \texttt{\{core, episodic, life, work, interest\}}, then call \texttt{finish\_memory\_update()}.

\textbf{Memory definitions.}
\begin{itemize}\itemsep0pt
\item \texttt{core}: WHO the user is and HOW to interact (identity, preferences, relationships, static reference data).
\item \texttt{episodic}: WHAT happened WHEN (time-anchored events and activities).
\item \texttt{life}: daily life, health, family, social, food, shopping.
\item \texttt{work}: professional or academic activities, projects, meetings, learning.
\item \texttt{interest}: hobbies, entertainment, leisure.
\end{itemize}

\textbf{Routing rules.}
\begin{itemize}\itemsep0pt
\item Include \texttt{episodic} for nearly all messages containing events.
\item Include \texttt{life} / \texttt{work} / \texttt{interest} only when relevant content is present.
\item Include \texttt{core} only when the message reveals new fundamental user information.
\item When uncertain between scenes, prefer including more rather than missing relevant content.
\end{itemize}

\textbf{Example.} \\
Input: ``Went to the gym with my colleague after work.'' \\
Output: \texttt{trigger\_memory\_update([episodic, work, life])}
\end{promptbox}
\captionof{figure}{Routing prompt used by the Meta Memory Manager at storage time.}
\label{fig:prompt_meta}

\bigskip

\begin{promptbox}{Planner LLM}
\textbf{Role.} You are the retrieval planner of CreaMem. Given a user question, select which memory components to search and the keywords to use.

\textbf{Task.} Output a JSON object with two fields:
\begin{itemize}\itemsep0pt
\item \texttt{keywords}: 3 to 6 keywords extracted from the question, used for BM25 and embedding retrieval.
\item \texttt{memory components}: a subset of \texttt{\{episodic, life, work, interest\}} indicating which components to query. Include the entire Core Memory if you think answering the question requires the user's profile.
\end{itemize}

\textbf{Selection guidance.}
\begin{itemize}\itemsep0pt
\item Questions about specific events or timestamps $\rightarrow$ include \texttt{episodic}.
\item Questions about user traits, habits, or preferences $\rightarrow$ include the relevant Life Scene Memories.
\item Multi-aspect questions $\rightarrow$ select multiple components
\end{itemize}

\textbf{Fallback.} On parse failure or empty output, the system retrieves from all four components with the question itself as the keyword.

\textbf{Example.} \\
Question: ``When did the user go camping with family?'' \\
Output: \texttt{\{"keywords": ["camping", "family", "trip"], "components": ["episodic", "life"]\}}
\end{promptbox}
\captionof{figure}{Planner prompt that produces $(K, R)$ at retrieval time.}
\label{fig:prompt_planner}

\bigskip

\begin{promptbox}{Episodic Memory Extractor}
\textbf{Role.} You are the Episodic Memory Manager. You receive batched user messages with timestamps and extract distinct events as separate entries.

\textbf{Schema.} Each episodic entry is a tuple \texttt{(event\_type, summary, details, actor)} with timestamp inherited from the conversation.

\textbf{Extraction rules.}
\begin{itemize}\itemsep0pt
\item Identify all distinct events in the batch; each event becomes a separate entry.
\item Convert relative time references (``yesterday'', ``last week'') to absolute dates using the conversation timestamp.
\item Preserve proper nouns, numbers, titles, and concrete details verbatim.
\item Skip greetings, small talk, and content already present in existing entries.
\end{itemize}

\textbf{Tool calls.} Use \texttt{episodic\_memory\_insert(items=[...])} by default. Use \texttt{episodic\_memory\_merge} only to continue the exact same event within one turn. Make exactly one tool call per batch.

\textbf{Example.} \\
Input (timestamp 2023-07-06): ``We went camping at the beach last weekend, and I painted a sunset last year.'' \\
Output:
\begin{verbatim}
episodic_memory_insert([
  {event_type: activity,
   summary: "User went camping at the
            beach around July 1-2, 2023",
   details: "Camped at the beach with
             family on 2023-07-01 to 07-02",
   actor: user},
  {event_type: activity,
   summary: "User painted a sunset in 2022",
   details: "User painted a sunset in 2022",
   actor: user}
])
\end{verbatim}
\end{promptbox}
\captionof{figure}{Extraction prompt for the Episodic Memory.}
\label{fig:prompt_episodic}

\bigskip

\begin{promptbox}{Scene Memory Extractor (Life / Work / Interest)}
The three Life Scene Memories share a common extractor template. Only the scene definition (highlighted below) differs across instantiations.

\textbf{Role.} You are the $\{$\textbf{SCENE}$\}$ Memory Manager of CreaMem. You extract scene-specific traits from batched user messages.

\textbf{Scene definitions.}
\begin{itemize}\itemsep0pt
\item \texttt{life}: health, family, daily routines, food preferences, social interactions, shopping, medical.
\item \texttt{work}: current role, projects, professional skills, deadlines, learning, career goals.
\item \texttt{interest}: hobbies, entertainment, sports, music, movies, travel, creative pursuits.
\end{itemize}

\textbf{Schema.} Each scene entry is a tuple \texttt{(content, importance\_score)}, where \texttt{content} is a trait, and \texttt{importance\_score} $\in [0, 1]$.

\textbf{Extraction rules.}
\begin{itemize}\itemsep 0pt
\item Extract distilled trait inferences, not raw events.
\item Each distinct fact becomes a separate entry.
\item Preserve proper nouns, numbers, and concrete details verbatim.
\item Ignore content outside the assigned scene.
\end{itemize}

\textbf{Tool calls.} Use \texttt{$\{$scene$\}$\_memory\_insert(items=[...])} by default. Use \texttt{$\{$scene$\}$\_memory\_update} only when an existing entry's information has genuinely changed.

\textbf{Example (Life).} \\
Input: ``I'm single, moved from Sweden 4 years ago, and have 3 pets: Oliver (dog), Luna (cat), Bailey (dog).''\\
Output:
\begin{verbatim}
life_memory_insert([
  {content: "User is single",
   score: 0.7},
  {content: "User moved from Sweden in 2019",
   score: 0.8},
  {content: "User has 3 pets: Oliver (dog),
            Luna (cat), Bailey (dog)",
   score: 0.7}
])
\end{verbatim}
\end{promptbox}
\captionof{figure}{Shared extraction template used by the three Life Scene Memories. Only the scene definition varies across instantiations.}
\label{fig:prompt_scene}

\bigskip

\begin{promptbox}{QA Prompt}
You are an intelligent dialog bot. You will be shown History Dialogs. Please read, memorize, and understand the given Dialogs, then generate one concise, coherent and helpful response for the Question.

\medskip
\texttt{History Dialogs: \{retrieved\_texts\}} \\
\texttt{Question Date: \{question\_date\}} \\
\texttt{Question: \{question\}}
\end{promptbox}
\captionof{figure}{QA prompt used to generate responses, identical to that of MemGAS~\cite{memgas2025} (following \citealp{lu2023memochat, pan2025secom}).}
\label{fig:prompt_qa}

\bigskip

\begin{promptbox}{GPT-4o Judge Prompt}
I will give you a question, a reference answer, and a response from a model. Please answer \texttt{[[yes]]} if the response contains the reference answer. Otherwise, answer \texttt{[[no]]}. If the response is equivalent to the correct answer or contains all the intermediate steps to get the reference answer, you should also answer \texttt{[[yes]]}. If the response only contains a subset of the information required by the answer, answer \texttt{[[no]]}.

\medskip
\texttt{[User Question]} \\
\texttt{\{question\}} \\
\texttt{[The Start of Reference Answer]} \\
\texttt{\{answer\}} \\
\texttt{[The End of Reference Answer]} \\
\texttt{[The Start of Model's Response]} \\
\texttt{\{response\}} \\
\texttt{[The End of Model's Response]}

\medskip
Is the model response correct? Answer \texttt{[[yes]]} or \texttt{[[no]]} only.
\end{promptbox}
\captionof{figure}{GPT-4o judge prompt used to score response correctness against ground truth, identical to that of MemGAS~\cite{memgas2025}.}
\label{fig:prompt_judge}

\end{document}